\documentclass{dhnbpub}

\usepackage{listings}
\usepackage{bbding}

\usepackage{tikz}

\begin{document}

\copyrightyear{2025}
\copyrightclause{Copyright for this paper by its authors.
  Use permitted under Creative Commons License Attribution 4.0 International (CC BY 4.0).}

\conference{The 9th Digital Humanities in the Nordic and Baltic Countries Conference (DHNB 2026), March 9–13, 2026, Aarhus, Denmark}

\title{Molecules of a Story: Community Detection in PMI-weighted Narrative Networks}

\author[1,2,3]{Kasper Fyhn}[%
orcid=0009-0003-5240-3661,
email=kfyhn@cc.au.dk,
url=https://pure.au.dk/portal/da/persons/kfyhn@cc.au.dk/,
]
\cormark[1]

\author[1,2]{Rebekah Baglini}[%
orcid=0000-0002-2836-5867,
email=rbkh@cc.au.dk,
url=https://pure.au.dk/portal/da/persons/rbkh@cc.au.dk/,
]

\address[1]{Department of Linguistics and Cognitive Science, Aarhus University}
\address[2]{TEXT - Center for Contemporary Cultures of Text, Aarhus University}
\address[3]{Center for Humanities Computing, Aarhus University}

\cortext[1]{Corresponding author.}

\begin{abstract}
Automatically extracted narrative networks — graphs with entities as nodes and their relations as edges — have proven useful for revealing central narrative structures through salient entities and their connections \citep{tangherlini_automated_2020, labatut_extraction_2019}. But a narrative is more than those central structures that everything else revolves around.

This work is concerned with the \textit{everything else}: brief sub-plots, small clusters of descriptions, or associations between minor characters that go under the radar at the macro-level. We present an approach to unearth such peripheral structures. They involve rare entities with limited textual presence, overshadowed by dominant entities and lost among each other in the long tail of many but rare entities \citep{baayen_word_2001}. We leverage the known tendency of pointwise mutual information \citep[PMI,][]{church_word_1990} to inflate for rare events, turning its weakness into a strength by weighting edges with PMI to foreground peripheral entity configurations. Communities extracted from the resulting network are structural traces of underlying narrative elements.

We demonstrate the approach on \textit{The Lord of the Rings}. From measures of how concentrated or dispersed a community's activations are across the text, a typology emerges that reveals that peripheral structures form more than a single class: episodic passages, echoing long-distance connections, and recurring threads each surface as distinct configurations. The approach is conceptually simple and surfaces fine-grained narrative details that are lost in abundance, though its deliberate amplification of weak signals comes with inherent sensitivity — best understood as a lens for exploration rather than a robust extraction pipeline.

\end{abstract}

\begin{keywords}
  narrative networks \sep
  rare entities \sep
  community detection \sep
  natural language processing
\end{keywords}

\maketitle

\section{Introduction}
Automatically extracted narrative networks — graphs with entities as nodes and their relations as edges — reflect structures of a narrative \citep[e.g.,][]{holur_modelling_2021,tangherlini_automated_2020,labatut_extraction_2019}. Such networks have proven particularly useful in foregrounding central narrative structures of salient entities and their connections at a macro-level, typically by highlighting salient entities and their interactions through frequency-based or centrality-based measures.

However, such approaches tend to emphasize dominant elements of a narrative while providing limited access to less frequent or more localized configurations. Narratives contain a wide range of such configurations, including brief descriptive passages, minor interactions, and associations among infrequent entities. According to Zipf's law, the distribution of words or entities extracted from a corpus has a long tail of rare ones. On one hand, such rare entities are easily overshadowed by more salient entities. On the other hand, they are lost among each other in the long tail of many but rare entities \citep{baayen_word_2001} along with noise from extraction errors. They form a seemingly unstructured mass, yet there \textit{is} structure to be recovered in how entities belong together in few but specific configurations. As such, the problem is not about needles in a haystack, but about how the straws of the haystack fit together.

To recover this structure, we propose using pointwise mutual information \citep[PMI,][]{church_word_1990}, which has a known tendency to inflate for rare entities. Here, we turn its weakness to a strength. We leverage the inflation to foreground these structures by weighting edges with PMI and extracting communities from a network that thereby favors the rarities. The amplification of weak signals comes with inherent sensitivity — a property we return to in the discussion. The extracted communities represent very different parts of a narrative – who's, what's, where's, when's and why's \citep[cf.][]{piper_narrative_2021} – and can be considered structural traces of peripheral narrative elements. Whether eventful, descriptive or even lyrical, they pertain to the size, shape, and composition of smaller parts of narrative that computational studies can reveal \citep{piper_computational_2023}. As molecules of a larger narrative structure, they contribute texture and specificity to the whole.

We demonstrate the approach on \textit{The Lord of the Rings} trilogy as a complex narrative work whose richness in sub-plots, world-building, and peripheral narrative elements makes it a suitable test case for the extraction of peripheral structures. No benchmark exists for our task and the approach is at an exploratory stage. Evaluation is therefore qualitative: we examine whether extracted communities correspond to recognizable structures in the source text.

The contribution of this paper is twofold. Methodologically, it demonstrates how PMI weighting can be used to systematically foreground low-frequency co-occurrences in narrative networks and how community detection can organize these into analyzable structures. Analytically, it provides an initial characterization of variation among such structures based on their distribution across a text. The approach is intended as a complement to existing methods that focus on salient entities, offering a means of exploring structure in the long tail rather than replacing established techniques.

\section{Methods}
Narrative graphs model a text as a network where nodes represent entities mentioned in the text and edges represent relations between them. This is an entity-centric approach rather than an event-centric approach, where the temporal axis and relations between events are backgrounded to prioritize a view of entities and their relations. Since \citet{bearman_becoming_2000}, the approach has been applied in various settings and with different types of relations \citep{labatut_extraction_2019}. In this work, we use a co-occurrence graph: nodes are entities and an edge between two entities indicates that they co-occur within a limited text span.

\subsection{Graph construction}
We create a narrative graph from a text collection in three steps. The analytical focus is on PMI weighting and community detection rather than on sophisticated entity extraction, and the pipeline is kept accordingly simple. Coreference resolution is included as the linguistically principled default for entity-centric analysis.

\paragraph{Entity extraction} We use a SpaCy pipeline \citep{honnibal_spacy_2020} augmented with coreference resolution via fastcoref \citep{otmazgin_f-coref_2022} to extract entities in a broad sense — characters, locations, objects, and concepts. Coreference resolution links mentions to their antecedents, so that pronouns and other anaphoric expressions are captured as occurrences of the entities they refer to. Named Entity Recognition targets character names and locations; Noun Phrase Chunking extends coverage to objects and concepts. To limit generic nouns, noun phrases must consist of at least two tokens.

\paragraph{Entity resolution} Extracted entities are resolved through two heuristic steps. First, entities are mapped to a canonical form via lemmatization. Second, supergrams are mapped to their shortest subgram (such as \textit{the dark lord Sauron} to \textit{the dark lord}), provided the subgram is at least twice as frequent and at least two tokens long. These conditions balance resolving different mentions of the same entity against retaining weak signals from the rare entities that we target. Beyond resolution, we apply minimal filtering: single-occurrence entities are removed as likely extraction noise, but otherwise the foregrounding of meaningful entities is left to the subsequent community detection step.

\paragraph{Co-occurrence extraction} Co-occurrences are extracted using a sliding window of three sentences. Each pair of entities appearing within the same window is counted as one co-occurrence.

\subsection{PMI-weighted edges}

PMI is calculated for two entities $A$ and $B$ as in \citep{church_word_1990}:
\begin{displaymath}
    PMI(A, B) = \log \frac{P(A, B)}{P(A) P(B)}
\end{displaymath}

Entity occurrence probabilities are obtained through maximum likelihood estimation from corpus frequencies: $P(A)=f(A)/N$
where $N$ is the total number of entity mentions. Their joint probabilities are estimated by counting the number of times that $A$ and $B$ both appear in the co-occurrence window described above: $P(A,B)=f(A,B)/N$. A positive PMI indicates that two entities co-occur more than expected by chance; a negative PMI indicates the opposite.

\begin{figure}
    \centering
    \includegraphics[width=0.5\linewidth]{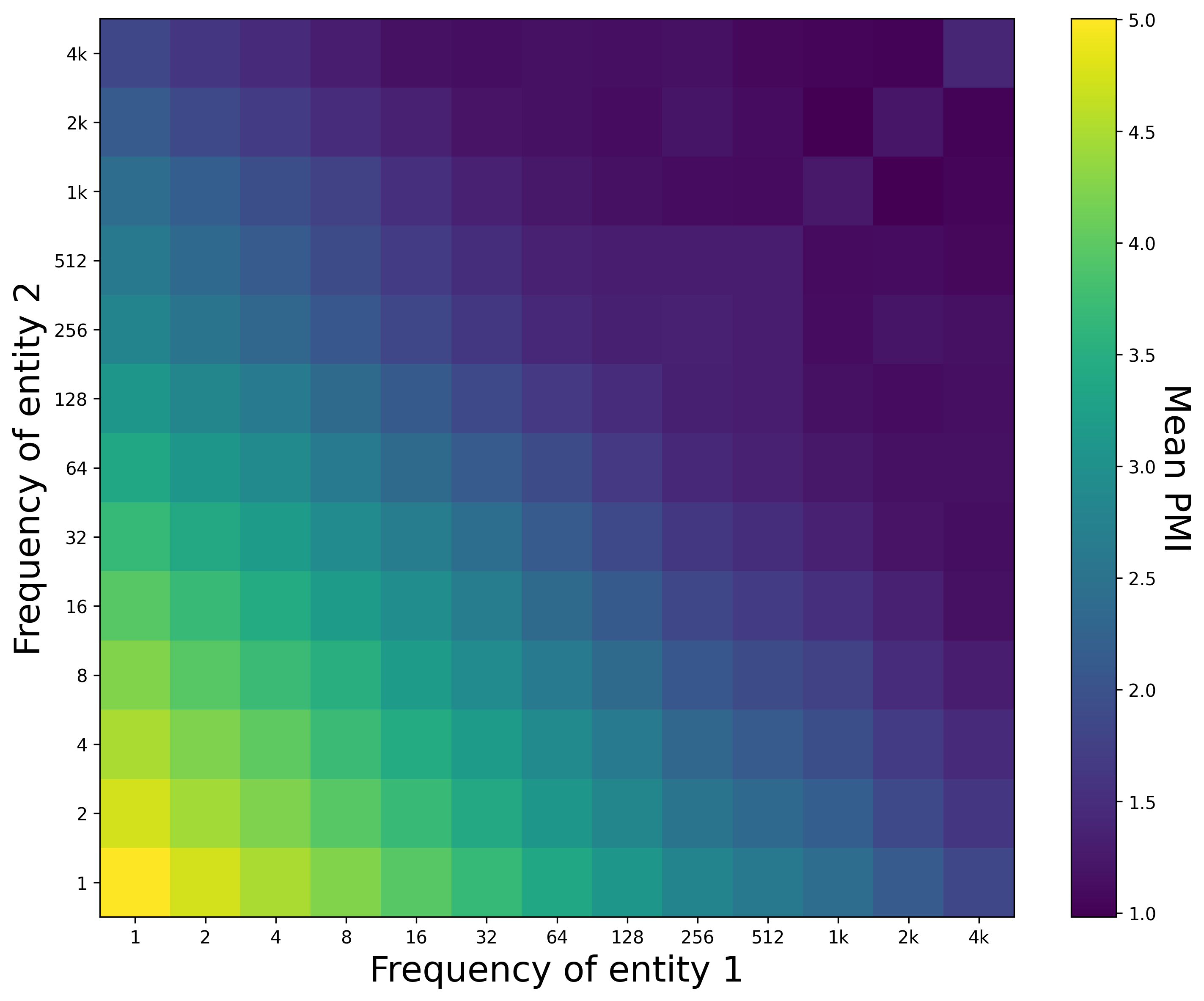}
    \caption{Mean PMI by entity frequency in \textit{The Lord of the Rings}. Pairs with low frequency entities generally have a much higher PMI than pairs with high frequency entities.}
    \label{fig:pmi-by-freq}
\end{figure}

PMI has a known tendency to inflate for rare events: if two entities are individually rare, even a single co-occurrence can yield a high PMI value. This is typically considered a limitation; \citet{church_word_1990} note that the measure becomes unstable below a frequency of 5. For our purposes, however, this property is central. The inflation foregrounds precisely the rare entity pairs that form the basis of the peripheral structures. Figure \ref{fig:pmi-by-freq} illustrates this with a heatmap of mean PMI by entity frequency brackets in \textit{The Lord of the Rings}. Low-frequency entity pairs have considerably higher PMI than high-frequency ones.

\subsection{Community detection}
Given a PMI-weighted graph that foregrounds connections between rare entities, we employ community detection to identify tightly connected subgraphs as candidates. We compare two methods that differ in two key properties: whether they allow overlapping communities, and whether PMI is treated as continuous weights or used for filtering. Both are applied using their implementations in the \textit{NetworkX} package \citep{hagberg_exploring_2008}.

The Louvain community detection algorithm \citep{blondel_fast_2008} is a hierarchical method that iteratively merges communities while optimizing for modularity. It partitions the graph: each entity belongs to exactly one community. In a narrative setting, this is a strong constraint, as entities naturally participate in multiple narrative configurations. On the other hand, Louvain supports continuous edge weights, operating directly on the PMI-weighted graph without requiring a threshold. Its key hyperparameter, \textit{resolution}, controls community granularity.

In k-clique percolation, a community is defined as "a union of all \textit{k}-cliques (complete subgraphs of size $k$) that can be reached from each other through a series of adjacent \textit{k}-cliques (where adjacency means sharing $k-1$ nodes)" \citep[][815]{palla_uncovering_2005}. Crucially, the method allows communities to overlap, reflecting the narrative reality that characters and locations participate in more than one subplot. The method has two hyperparameters: $k$, the minimum clique size, and a weight threshold $w$ that filters edges below a given PMI value. The weight threshold interacts with PMI weighting: filtering by $w$ removes weaker, more frequent connections, retaining the strong associations between rare entities.

\subsection{Measures of textual confinement}
\label{sec:spikes-and-spread}
A community can be activated at certain points over the course of a text. An activation means that some of the entities of a community co-occur within a text passage — the same co-occurrences that back the edges of the graph. We refer to a document that contains at least one such activation for a given community as a \textit{spike}. For \textit{The Lord of the Rings}, we take sections delimited by double line breaks as documents.

Given an ordering of documents, we characterize a community by the \textit{spread} of its spikes: the distance from first to last activation, measured in documents. A community with all its spikes in consecutive or near-consecutive documents has low spread; one whose spikes are separated by long stretches of text has high spread. Note that the number of spikes is bounded by spread: a community cannot be activated in more documents than it spans, so the space of possible configurations is triangular rather than rectangular.

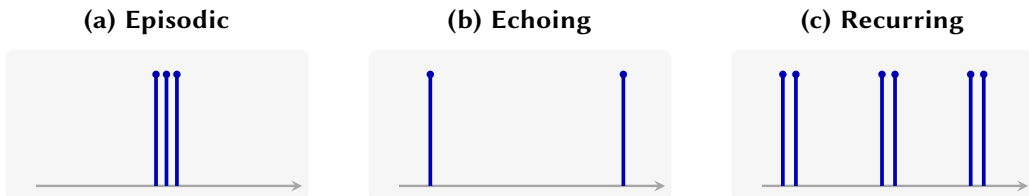
\begin{figure}
    \centering
    \begin{tikzpicture}
    \def\pw{4.0}
    \def\ph{2.0}
    \def\hgap{0.8}
    \def\ybase{0.1}
    
    \tikzset{
      myaxis/.style={->, >=stealth, thick, gray!70},
      spike/.style={blue!70!black, line width=1.4pt},
      baseline/.style={gray!50, line width=0.9pt},
    }
    
    \newcommand{\drawpanel}[3]{%
      \fill[gray!7, rounded corners=3pt] (#1, #2) rectangle (#1+\pw, #2+\ph);
      \draw[baseline] (#1+0.4, #2+\ybase*\ph) -- (#1+\pw-0.15, #2+\ybase*\ph);
      \draw[myaxis] (#1+0.4, #2+\ybase*\ph) -- (#1+\pw-0.08, #2+\ybase*\ph);
      \node[font=\small\bfseries, anchor=south, text width=\pw cm, align=center] at (#1+\pw/2, #2+\ph+0.08) {#3};
    }
    
    \newcommand{\spike}[3]{%
      \pgfmathsetmacro{\sx}{#1 + 0.4 + #3*(\pw-0.55)}
      \pgfmathsetmacro{\sybot}{#2 + \ybase*\ph}
      \pgfmathsetmacro{\sytop}{#2 + \ybase*\ph + 0.82*\ph*(1-\ybase)}
      \draw[spike] (\sx, \sybot) -- (\sx, \sytop);
      \fill[blue!70!black] (\sx, \sytop) circle (1.4pt);
    }
    
    %% Panel 1: Episodic — concentrated spikes, low spread
    \pgfmathsetmacro{\pA}{0}
    \drawpanel{\pA}{0}{(a) Episodic}
    \spike{\pA}{0}{0.46}
    \spike{\pA}{0}{0.50}
    \spike{\pA}{0}{0.54}
    
    %% Panel 2: Echoing — few spikes, high spread
    \pgfmathsetmacro{\pB}{\pw+\hgap}
    \drawpanel{\pB}{0}{(b) Echoing}
    \spike{\pB}{0}{0.12}
    \spike{\pB}{0}{0.86}
    
    %% Panel 3: Recurring — multiple spikes in clusters, high spread
    \pgfmathsetmacro{\pC}{2*\pw+2*\hgap}
    \drawpanel{\pC}{0}{(c) Recurring}
    % cluster 1
    \spike{\pC}{0}{0.08}
    \spike{\pC}{0}{0.13}
    % cluster 2
    \spike{\pC}{0}{0.46}
    \spike{\pC}{0}{0.51}
    % cluster 3
    \spike{\pC}{0}{0.80}
    \spike{\pC}{0}{0.85}
    
    \end{tikzpicture}
    \caption{Conceptual illustration of the three prototypical peripheral structures. Spikes (vertical lines) represent documents where part of a community is activated. The horizontal axis represents the sequential or temporal ordering of documents. (a)~Episodic structures have concentrated activations within a limited span. (b)~Echoing structures have few activations at distant points. (c)~Recurring structures have multiple activations spread across the text.}
    \label{fig:spikes-and-spread}
\end{figure}

We use three reference points within this space to orient our qualitative analysis (Figure \ref{fig:spikes-and-spread}): \textit{episodic} (concentrated activations within a limited span), \textit{echoing} (few activations at distant points), and \textit{recurring} (multiple activations spread across the text). These are prototypes in a continuous space rather than clear-cut categories. What they correspond to in terms of narrative function is an empirical question that we address in the analysis.

\section{Results}
\subsection{Comparison of community detection algorithms}
The two methods can produce quite different communities depending on hyperparameter settings. We selected hyperparameters that yield comparable community sizes: $k=4$ and $w=2$ for k-clique percolation and $resolution = 35$ for Louvain. As noted by \citet{palla_uncovering_2005}, k-cliques often result in at least one dominant community that is considerably bigger than the rest. We also observed these and discard them from the analysis. Descriptive statistics are presented in Table \ref{tab:comm-numbers}.

\begin{table}
    \centering
    \caption{Descriptive statistics of the number and sizes of communities extracted from \textit{The Lord of the Rings}}
    \begin{tabular}{cccccccc}\toprule
         &  No. of communities&   \multicolumn{6}{c}{Size}\\
 & & Mean&  Min&Q1&Median & Q3&Max\\\midrule
         k-clique&  1193 &  7.22 &   4&4&6  & 8&93\\
         Louvain&  797 &  7.0 &   2&4&6  & 9&24\\ \bottomrule
    \end{tabular}
    \label{tab:comm-numbers}
\end{table}

At comparable community sizes, k-clique produces considerable more communities than Louvain. This is expected because k-clique allows overlapping communities, meaning that the same entities can appear in multiple communities. On average, an entity belongs to 1.9 k-clique communities.

From manual inspection, both methods produce communities that correspond to recognizable narrative elements. However, both also exhibit a shared characteristic: the extracted communities are often fragments of a broader narrative configuration. Consider the following community: 
\begin{quote}
\textbf{Community}

\textit{a faint hiss, deep shade, black holes, the shapes}

\textbf{Context}

So black were they that they seemed like \textbf{black holes} in the \textbf{deep shade} behind them. Frodo thought that he heard \textbf{a faint hiss} as of venomous breath and felt a thin piercing chill. Then \textbf{the shapes} slowly advanced. Terror overcame Pippin and Merry, and they threw themselves flat on the ground. Sam shrank to Frodo's side.
\end{quote}

The passage is, in short, a part of the events that unfold at Weathertop where the Ringwraiths pursue, find and approach the four hobbits. The entities in the community pertain to the description of the Ringwraiths. Not all entities involved in the scene constitute the community, but a coherent subset of them does. This emphasizes how the community structure \textit{reflects} a peripheral element of a narrative rather than fully constituting it.

Given comparable quality, we adopt k-clique percolation as the primary method, but the reported findings generally hold for Louvain communities as well. K-clique's allowance of overlapping communities better reflects the narrative reality that characters and locations participate in more than one configuration. That it produces many more communities at comparable sizes is a consequence of this property.

\subsection{Typology}
Running k-clique percolation on the PMI-weighted narrative network of \textit{The Lord of the Rings} yields 1193 communities. Roughly three quarters are single-spike communities, activated in only one section, and therefore fall under the episodic type. Of the remaining multi-spike quarter, roughly a quarter are episodic (with multiple spikes in close proximity), roughly half are echoing, and the last quarter are recurring. Representative examples of all types as well as their subtypes can be seen in Table \ref{tab:lotr-examples}.

\begin{table}[htbp]
\centering
\caption{Representative examples of the community typology by spread and spikes}
\begin{tabular}{llp{2cm}p{5.5cm}p{4cm}@{}}
\toprule
 \textbf{Spread} & \textbf{Spikes} & \textbf{Function} & \textbf{Community} & \textbf{Gloss} \\\midrule
 \multicolumn{5}{c}{\textbf{\textit{Episodic}}}\\ 
 Low & Few & Description 
  & the stairway, the high regions, the main pass, the great ravine, the torment& Description of the climb and stairway in the mountains of the Morgul Vale into Mordor.\\
 Low & Few & Event 
  & your lordship, the herb-master, kingsfoil, latter days, old names& Aragorn's request for kingsfoil in the Houses of Healing, with which the herb-master returns two sections later. \\
 Low & Multiple& Description (sustained)& great height, a great circle, fosse, soft shadow, mallorn-trees, its brink, Green walls, ten miles, golden elanor, the white bridge, green hill, Their height& Description of areas and landmarks in Lothlorien, spanning multiple sections with short gaps between. \\
 \multicolumn{5}{c}{\textbf{\textit{Echoing}}}\\ 

 High & Few & Poem variation 
  & A new road, a secret gate, the hidden paths, Apple, the eastern sky, his memory& Distinct versions of a song sung by Frodo in the same geographical location early in book one and late in book three.\\
 High & Few & Textual echo& my fear, my doubts, sudden fear, the hobbit's ring& Gandalf says early in book one that his doubts slept uneasily regarding Bilbo's magical ring. Halfway through the book, Gandalf says how his doubts awakened.\\
 High& Few& Specific reference& the Old Took, the furniture, Tuckborough, Smials, our legs&An echoing reference to ``The Smials" in Tuckborough, ancestral home of the Tooks. \\
 \multicolumn{5}{c}{\textbf{\textit{Recurring}}}\\ 
 High & Multiple & Lore thread& the making, Ered Nimrais, Mount Mindolluin, their end, the highlands, Rammas Echor, a hardy folk, Belfalas, the herdsmen, Long slopes, Emyn Arnen, great labour, the Dark Years, the out-walls, the very brink, the quays, the southern fiefs, the townlands, grey eyes, a sudden end, Harlond, ten leagues, wide lands, the eastern shadow, Lossarnach, Lebennin, Green, their blood, the mountains' feet, Prince Imrahil& Descriptions of and references to fiefdoms and landmarks in the land of Gondor and around.\\
 High & Multiple & Recurring thread& black stone, their king, the terror, secret places, other men, Four days, the great ride, the Shadow Host, the last king, the forgotten people, the wrath, the Dark Years, Erech, the Men of the Mountains, Your oath, the Hill of Erech, the oath& The legend of the dead men of Dunharrow (oathbreakers resting in the mountains), how they are mustered for the war and relieved of their curse.\\

 High& Multiple& Recollection of threads& a little land, a foreboding,
his counsels, Tales, the end
of June, the southern bor-
ders, the Tale of the Ring,
Greenway, a chill&Gandalf recollects events at the
Council of Elrond that tie back to
short passages in the preceding half
of book one.\\ \bottomrule
\end{tabular}
\label{tab:lotr-examples}
\end{table}

Episodic structures are concentrated activations within a limited span. These are generally idiosyncratic in terms of entities which is expected from an information-theoretic perspective. Rare entities co-occur with few others, and small clusters in proximity will surface as communities with high internal PMI. What they correspond to varies: short descriptions, events, songs, or fragments of dialogue. As we saw with the example in the previous section, they can be interwoven with or themselves interweave other sub-narratives. A subset of episodic structures are sustained across multiple consecutive sections, e.g., concentrated sub-plots or scenic descriptions. These are few but distinct, and the more sustained they are, the less they are favored by PMI, since their entities will be more frequent.

Echoing structures (roughly a quarter) form a distinct class in which an idiosyncratic configuration reappears at a distant point in the text: a poem or song revisited, an event or quote repeated or referenced, or a specific (lore) reference between two or few entities. These long-distance connections are particularly interesting in a hybrid close-distant reading setting, where the computational method surfaces connections that a human reader may struggle to pick up across hundreds of pages. Below is an example where a vision by Sam in the Mirror of Galadriel in book one is realized by the end of book two.

\begin{quote}
\textbf{Community}

\textit{a winding stair, a dark veil, pale faces, a great dark cliff}

\textbf{Context 1}

As if \textbf{a dark veil} had been withdrawn, the Mirror grew grey, and then clear. There was sun shining, and the branches of trees were waving and tossing in the wind. But before Sam could make up his mind what it was that he saw, the light faded; and now he thought he saw Frodo with \textbf{a pale face} lying fast asleep under \textbf{a great dark cliff}. Then he seemed to see himself going along a dim passage, and climbing \textbf{an endless winding stair}.

\textbf{Context 2}

And suddenly he saw that he was in the picture that was revealed to him in the mirror of Galadriel in Lórien: Frodo with \textbf{pale face} lying fast asleep under \textbf{a great dark cliff}. Or fast asleep he had thought then. `He's dead! ' he said. '

\end{quote}

Recurring structures span much of the work and show internal variation. A small tail of many-spiked communities tends to be diluted by generic nouns, such as pairs like \textit{left side} and \textit{right side} that co-occur frequently but are questionable as meaningful narrative elements, or by aggregated connections between entities that do not form a meaningful cluster. Among the meaningful communities, some function as lore threads reflecting recurring geographic or world-building references, others track recurring sub-plots across distant parts of the narrative, and a third pattern emerges where a later passage gathers threads dispersed across earlier sections into a retrospective account.

\section{Discussion and future work}
The central idea of our work is to enhance weak signals by treating the statistical properties that make rare entities unreliable in conventional analyses as a resource rather than a nuisance. The approach is conceptually simple: weight a narrative network with PMI, extract communities from it, and characterize the output by simple structural measures. PMI amplifies rare co-occurrences, and community detection organizes those amplified signals into interpretable clusters. Whether or not every resulting community is meaningful, the principle — that structure exists in the long tail and can be made visible through deliberate amplification — is, we believe, broadly applicable beyond the present setting.

The typology that emerges from characterizing the extracted communities by their spikes and spread reveals that peripheral narrative structures form more than a single class. The bulk are single-spike communities from idiosyncratic passages that are introduced and wrapped up within a single section. One might question their value in a macro-perspective, but their sheer prevalence may hint at structural properties of narrative, namely that stories constantly introduce and resolve small elements alongside their central progression. The multi-spike communities, in turn, reveal how peripheral elements can also echo or recur across a narrative. This is particularly notable in how they surface weak-signal connections between elements across long textual distances. Such connections are easy to overlook and invisible to computational methods that foreground salience.

More broadly, the method can be regarded as a facilitator of further analysis. It does not replace interpretive work. The extracted communities are candidates, often even fragmentary ones, not conclusions. But it surfaces specific entity configurations, and these trace back to passages that can then be examined. This aligns with an idea of complementary contributions of computational and traditional literary analysis, where quantitative methods help recognize and surface patterns at scale \citep{long_literary_2016,jockers_macroanalysis_2013}. The long tail of rare entities can at first appear as an unstructured mass, but our method recovers non-random structures within it.

The described method deliberately leverages the instability of PMI at low frequencies \citep{church_word_1990}, which is a direct consequence of the unreliability of MLE probability estimates for rare events \citep{baayen_word_2001, dunning_accurate_1993}. The extracted communities depend on PMI values, which depend on marginal and joint probability estimates, which depend on entity counts. Operating in low-frequency territory, even small differences in individual entity counts propagate through each of these steps. This makes the method highly susceptible to extraction errors: at the level of rare entities, a single false positive or negative can distort PMI associations drastically, unlike macro-level patterns where signal strength masks such errors. Similarly, any changes to count input, such as coreference resolution or different entity resolution strategies, affect PMI values considerably. The method is also sensitive to hyperparameter choices. $k$ and the weight threshold for k-clique percolation can shift individual community assignments, though we observed that some communities are more stable under such variation than others — an observation worth formalizing in future work. This inescapable sensitivity illustrates that the method is best understood as a bottom-up lens for surfacing structure in the long tail of rare entities, rather than a robust pipeline for extracting predefined latent structures.

On the methodological side, other association measures such as log-likelihood \citep{dunning_accurate_1993}, which is more robust at low frequencies, may complement or improve the approach, and community detection algorithms that handle overlap differently offer a natural point of comparison. Empirically, application to other literary works and registers is a natural extension: more minimalist works with fewer characters or sparser descriptions may yield different results, and whether this reflects fewer peripheral structures or a different relationship between peripheral and central elements remains open. The structures recovered here are also characterized bottom-up; stronger grounding in narrative theory \citep{piper_narrative_2021} would complement that.

We have demonstrated that quantitative methods need not favor only salient signals. By treating PMI's inflation for rare events as a strength rather than a weakness and applying community detection, we surface non-random structure among rare entities — peripheral structures that trace back to recognizable parts of the source text, and that can otherwise be lost in abundance.

\begin{acknowledgments}
This work was supported by Danish Data Science Academy, which is funded by the Novo Nordisk Foundation (NNF21SA0069429) and VILLUM FONDEN (40516). This work was partially supported by the Danish National Research Foundation (DNRF193) through TEXT: Center for Contemporary Cultures of Text, Aarhus University.
\end{acknowledgments}

\printbibliography

\end{document}